\documentclass[10pt, conference, onecolumn]{IEEEtran}
\IEEEoverridecommandlockouts

\usepackage{algorithm}
\usepackage[noend]{algpseudocode} 
\usepackage{algorithmicx}

\usepackage{amsmath,amssymb,amsfonts}
\usepackage{float}
\usepackage{multicol}
\usepackage{lipsum} 
\usepackage{graphicx}
\usepackage{textcomp}
\usepackage{xcolor}
\usepackage{subcaption}
\usepackage{breakurl}
\usepackage{xurl}
\usepackage{booktabs}  
\usepackage{longtable} 
\usepackage{array}     
\usepackage{verbatim}

\def\BibTeX{{\rm B\kern-.05em{\sc i\kern-.025em b}\kern-.08em
    T\kern-.1667em\lower.7ex\hbox{E}\kern-.125emX}}

\begin{document}
\title{Reinforcement Learning-Driven Edge Management for Reliable Multi-view 3D Reconstruction\thanks{This material is based upon work supported by the National Science Foundation (NSF) under Award Number CNS-1943338}}

\author{
Motahare Mounesan, Sourya Saha, Houchao Gan, Md. Nurul Absur, Saptarshi Debroy\\
City University of New York\\ 
Emails: \textit{\{mmounesan,ssaha2,hgan,mabsur\}@gradcenter.cuny.edu, saptarshi.debroy@hunter.cuny.edu}}

\maketitle

\begin{abstract}
Real-time multi-view 3D reconstruction is a mission-critical application for key edge-native use cases, such as fire rescue, where timely and accurate 3D scene modeling enables situational awareness and informed decision-making. However, the dynamic and unpredictable nature of edge resource availability introduces disruptions, such as degraded image quality, unstable network links, and fluctuating server loads, which challenge the reliability of the reconstruction pipeline. In this work, we present a reinforcement learning (RL)-based edge resource management framework for reliable 3D reconstruction to ensure high quality reconstruction within a reasonable amount of time, despite the system operating under a resource-constrained and disruption-prone environment. In particular, the framework adopts two cooperative Q-learning agents, one for camera selection and one for server selection, both of which operate entirely online, learning policies through interactions with the edge environment. To support learning under realistic constraints and evaluate system performance, we implement a distributed testbed comprising lab-hosted end devices and FABRIC infrastructure-hosted edge servers to emulate smart city edge infrastructure under realistic disruption scenarios. Results show that the proposed framework improves application reliability by effectively balancing end-to-end latency and reconstruction quality in dynamic environments.
\end{abstract}

\thispagestyle{empty}


\section{Introduction}
Multi-view 3D reconstruction of unknown scenes is becoming a foundational capability for mission-critical smart city applications, such as emergency response and public safety, providing situational awareness for rapid, informed decision making~\cite{10.1145/2459236.2459271,10.1145/3583740.3630267}. This technique aggregates images from multiple viewpoints to generate 3D representations, typically as point clouds or meshes. Reconstructing 3D scenes from overlapping views requires substantial computational resources. While data centers can easily handle such workloads, smart city deployments rely on city-scale edge servers for lower latency, cost efficiency, and data privacy. While these edge servers are geographically closer and suitable for `near real-time' processing, they lack the capacity of remote data centers. Delivering timely results thus necessitates intelligent resource management to balance conflicting performance requirements.

The already tricky proposition of edge management for multi-view 3D reconstruction becomes even more challenging in mission-critical scenarios due to its susceptibility to disruptions and fluctuations. This comes from three factors: i) the nature of the reconstruction process, ii) the volatile environments of the underlying use cases, and iii) the resource constraints of smart city edge ecosystems. Unlike simpler video processing applications that operate on independent frame streams~\cite{effect,effectDNN, Shima,Manal}, 3D reconstruction requires high-quality, spatially and temporally aligned images from multiple viewpoints. When such data is degraded or partially missing, the reconstruction results can be compromised both quantitatively (e.g., incomplete surfaces and missing objects) and qualitatively (e.g., geometric distortions). This vulnerability is exacerbated in dynamic, high-risk settings, such as fire scenes, disaster zones, or urban deployments, where sensing and communication infrastructure is prone to disruption. Further, the inherently volatile nature of smart city infrastructure can also cause disruptions, including unstable network condition, limited edge connectivity, processing delays, and resource contention. Although these issues mainly affect end-to-end processing latency, degraded or unavailable input images cannot be retransmitted or recovered within the tight latency constraints of rapidly evolving environments.

Reducing end-to-end latency of multi-view 3D reconstruction pipelines~\cite{moulon2016openmvg, openmvs2020,yao2018mvsnet} without significantly degrading reconstruction quality is thus a challenging task, especially in smart city edge deployments where network conditions and camera availability fluctuate, timing constraints are strict, and the true system state is only partially observable~\cite{10.1145/3583740.3630267,zhang2022end}. Achieving consistent performance across these varying conditions requires a {\em reliable} decision-making process, one that adapts to disruptions while ensuring reconstruction results remain timely and accurate. Traditional heuristic-based approaches for edge management rely primarily on static environment and prior knowledge of system parameters — assumptions that are unrealistic in such dynamic environments. Consequently, their static design prevents them from ensuring reliable performance. Reinforcement learning (RL)-based decision-making, which enables the system to learn effective policies {\em online} without full knowledge of the environment in advance, can offer robust and adaptive solutions for managing the latency–quality trade-off in mission-critical applications.

In this paper, we focus on two critical decisions: selecting camera subsets, affected by visual or network disruptions, and choosing edge servers, influenced by computational load and network conditions. We develop an RL-based framework that enables the system to make these decisions adaptively. We employ two cooperative Q-learning agents, one for camera selection and one for server selection, learning adaptive policies that improve responsiveness and reliability under unpredictable disruptions. The agents learn entirely online during system operation, thus requiring practical implementation to expose the agents to representative system dynamics. Thus, we implement and evaluate our framework on a realistic testbed composed of resource-constrained end devices in the lab, along with FABRIC network and server infrastructure that emulates network and edge-server components, operating under diverse simulated disruption scenarios. 
Results demonstrate that the camera selection strategy achieves up to 15\% higher reliability, while the adaptive server selection strategy achieves up to a 50\% improvement over baselines. Overall, these findings highlight the potential of RL to enable robust, disruption-aware decision making in real-time edge systems. 



The remainder of this paper is organized as follows. Section~\ref{relatedworks} reviews the background and related work. Section~\ref{systemmodel} describes the system model and details the proposed reinforcement learning–based framework for camera and server selection. Section~\ref{evaluation} presents our testbed setup, evaluation methodology, and experimental results under various disruption scenarios. Finally, Section~\ref{sec:conclusions} summarizes the key findings.

\section{Background and related works}\label{relatedworks}
Here, we present an overview of multi-view 3D reconstruction and discuss related works. 

\subsection{Multi-view 3D reconstruction}
Multi-view 3D reconstruction generates 3D representations of objects or scenes from RGB images or occasionally sketches. It uses images captured from multiple perspectives, either as sequential frames or sparse views. Outputs include voxel grids, point clouds, and meshes~\cite{10301359}, with point clouds and meshes particularly versatile, scalable, and accurate. Geometry-based reconstruction follows a two-step pipeline: Structure-from-Motion (SfM) and Multi-View Stereo (MVS). OpenMVG/openMVS~\cite{moulon2016openmvg,openmvs2020} is a popular library implementing this pipeline. Deep learning methods estimate object geometry using prior knowledge~\cite{samavati2023deep, han2019image}, but often  yield lower accuracy and generalization than geometry-based approaches. \emph{In this work, we adopt the geometry-based pipeline for its robustness and reconstruction quality.}


\subsection{Evaluating reconstruction quality}
A key objective of latency-optimized edge management for 3D reconstruction is ensuring high-quality outputs, which requires reliable techniques to evaluate generated point clouds. Quality can be assessed via three main methods: Geometry-Oriented~\cite{Lin_2023_ICCV}, Color-Oriented~\cite{9123089}, and Combined~\cite{9450013}, all of which rely on {\em ground-truth} point clouds. However, ground truth for unknown 3D scenes in a smart city environment is typically unavailable. In its absence, point cloud quality must be evaluated using no-reference methods, including traditional~\cite{9506329} and learning-based approaches~\cite{liu2021pqa}. Traditional methods assess internal geometric consistency or appearance cues derived from projections, while learning-based methods predict perceptual metrics such as the Mean Opinion Score (MOS) using neural networks trained on annotated datasets. \emph{Since creating and managing large annotated datasets for city-scale indoor/outdoor scenes is impractical, we employ a latency-efficient geometry-based technique leveraging known intrinsic and extrinsic camera parameters (focal length, distortion, pose) to produce perspective-aligned images that preserve geometric structure and visual appearance. This generates high-fidelity projections that enable robust quality assessments even without ground truth.}

\subsection{Reinforcement Learning for edge resource management}
Edge resource management for complex, time-sensitive applications such as 3D reconstruction is challenging. Reinforcement learning has emerged as a promising approach for sequential decision-making in dynamic edge environments~\cite{VEC, edgerl, inferedge}. RL has been applied to optimize resource allocation for 3D pose reconstruction in robotic minimally invasive surgery~\cite{10684243}, manage edge server orchestration for augmented reality workloads~\cite{10293868}, and support adaptive wireless access in emergency response~\cite{9382256}. DRL-based frameworks like DRJOA~\cite{Chen2023} jointly optimize task offloading and wireless resources in MEC scenarios. {\em However, many RL-based methods rely on offline training or assume relatively static environments. Real-time, disruption-prone, resource-constrained settings like ours require online learning and rapid adaptation. Our work addresses this gap.}


\subsection{Managing disruptions in distributed systems}
Disruption management in distributed systems is becoming a critical area of research aimed at improving system reliability and security. Recent studies explore various aspects of this challenge. For example, the ReSeT system \cite{8685918} focuses on reducing service disruption by optimizing handover times for mobile nodes. 
Work, such as \cite{8643392} addresses latency in real-world applications by proposing a two-phase scheduling strategy to mitigate disruptions in edge environments. Additionally, \cite{absur2024posterreliable3dreconstruction} explores spatio-temporal disruption patterns that can impact video processing quality and applies a portfolio theory-inspired approach to model these disruptions. Video processing in edge systems is exposed to different types of disruptions beyond general system reliability. These include environmental and infrastructure constraints such as obstructions, occlusions \cite{He:20}, and mission-critical disruptions \cite{s23073529} that directly affect image quality and coverage. \emph{Unlike these, in our work, we aim to explore a combination of factors, including environmental and infrastructure constraints, alongside coverage, accuracy, and computational complexity. We propose a resource selection strategy that is specifically tailored to address the unique challenges of complex 3D video processing tasks to ensure high-quality reconstruction.}

\section{Problem Formulation and Solution Strategy}
\label{systemmodel}
Here, we discuss the proposed system model, RL based problem formulation, and Q-learning based solution strategy.

\vspace{-4pt}
\subsection{System model}
\label{sec:system model}
We consider a smart city environment running a 3D reconstruction application using city-hosted edge servers. The key components are: a set of edge servers for computation, a fleet of camera-equipped end devices (e.g., drones, robots) for scene capture, and a central edge controller or resource manager for overall resource allocation.


\noindent \underline{\textit{Edge servers ($\mathcal{E}$)} --} We consider a set of $m$ smart city edge servers, each responsible for performing computationally intensive 3D reconstruction tasks. These servers operate under dynamically changing workloads due to processing demands from a variety of smart city applications running simultaneously. The available computational capacity and processing latency of each server fluctuate over time, and are not directly controlled by the controller.\\
\noindent \underline{\textit{Camera-equipped end devices ($\mathcal{C}$)} --} We consider a fleet of $n$ end devices (e.g., drones, robots), each equipped with a high-resolution camera of identical hardware specifications but capturing the scene from different viewpoints. These devices continuously stream visual data to the edge system. However, environmental factors such as smoke, fog, or fire may degrade the quality of captured images, while wireless communication disruptions can prevent timely delivery of images to the controller. Both types of disruptions affect the availability and usefulness of each camera's data during processing runtime.

\noindent \underline{\textit{Controller} --}
We assume a centralized decision-making module operating at the network edge responsible for overall resource management of the entire smart city environment. In the context of the 3D reconstruction use case which serves as the problem at hand, the controller's objective is to make latency- and quality-aware decisions, adapting to disruptions in sensing, communication, and computation. In this context, we reiterate that such systems operate in a disruption-prone environment, where among issues,  unpredictable fluctuations in sensing quality, communication reliability, and computation capacity impact the {\em end-to-end latency and reconstruction quality balance} the most. 

While the system model can accommodate multiple resource allocations (e.g., network paths, wireless channels), this work focuses on camera selection and server assignment under dynamic conditions. The primary disruptions either block data from a camera, reducing reconstruction quality, or overload a server, increasing latency. We study the controller’s ability to adaptively select cameras and servers at each timestep in response to the evolving system state.

\vspace{-4pt}
\subsection{Defining 3D reconstruction reliability}
\label{sec:reliability}
A 3D reconstruction application is considered reliable if it consistently produces usable reconstructions within a predefined near real-time latency budget under dynamic conditions. Formally, we define the reliability ($\mathbb{R}_t$) at timestep $t$ as a binary variable: TRUE if both quality and latency constraints are satisfied, and FALSE otherwise. In other words, a camera and server selection is reliable if it meets the minimum quality threshold and maximum latency constraint. Thus:

\vspace{-10pt}
\begin{equation}
    \mathbb{R}_t = 
    \begin{cases}
    1, & \text{if } Q_t \geq \Theta \text{ and } L_t\leq \Phi \\
    0, & \text{otherwise}
    \end{cases}
    \label{eq:reliability}
    \vspace{-6pt}
\end{equation}

\noindent where:
\begin{itemize}
    \item \(Q_t\) denotes the point cloud quality score at timestep \(t\),
    \item \(L_t\) denotes the total end-to-end latency which includes transmission and processing (i.e., reconstruction) delays,
    \item \(\Theta\) is the minimum acceptable reconstruction quality,
    \item \(\Phi\) is the maximum allowable latency.
\end{itemize}

Long-term reliability is the proportion of timesteps with $\mathbb{R}_t = 1$, directly reflecting task success under predefined quality and latency constraints and supporting real-time evaluation.


\vspace{-4pt}
\subsection{RL-based problem formulation}

As reliability is defined as satisfaction of two constraints: (i) reconstruction quality must remain above a minimum threshold, and (ii) reconstruction latency must not exceed a predefined upper bound, the problem is formulated as a constrained optimization task to select camera subsets and edge servers that maximize reliability under operational constraints.


We model the camera and server selection problem as a Markov Decision Process (MDP) to enable RL-based optimization under uncertain and variable environmental conditions. While some system components are partially observable, the MDP framework allows decision-making as a sequence of actions guided by measurable outcomes such as reconstruction quality and latency. This supports policies that adapt over time and generalize across deployment scenarios where handcrafted heuristics fail and prior learning is infeasible. To manage complexity and enable efficient online learning, we decompose the process into two cooperative agents: one for camera selection and one for server selection. {\em This separation is motivated by the large combinatorial space of camera selection, which complicates online adaptation if coupled with server decisions. Server selection depends on broader system context, including chosen cameras and projected resource load. Decoupling these roles allows each agent to optimize its subproblem independently with tailored rewards.}

\noindent \underline{\textit{Camera selection} --}
The camera selection task is formulated as a \textit{stateless Markov Decision Process (MDP)}. To support real-time inference and reduce model complexity, the agent operates without an explicit state representation. It selects actions based solely on its learned policy, without access to dynamic environment variables such as camera availability, quality degradation, or network conditions. As a result, the environment is treated as stateless from the agent's perspective, and learning is driven purely by interaction and feedback.

At each timestep \( t \), the agent selects a subset of cameras from a pool of \( N \) candidates, based on the current set of valid combinations provided by the environment. The action space \( \mathcal{A}_{\mathcal{C}} \) is defined as:

\vspace{-10pt}
\begin{equation} 
    \mathcal{A}_{\mathcal{C}} = \left\{ \mathbf{a} \in \{0,1\}^N \mid k_{\text{min}} \leq \|\mathbf{a}\|_1 \leq k_{\text{max}} \right\}
    \vspace{-6pt}
\end{equation}

where each binary vector \( \mathbf{a} \) encodes which cameras are selected (\(1\)) or not (\(0\)), and the \( \ell_1 \) norm constrains the number of selected cameras within predefined bounds.

The reward received by the camera agent at each timestep \( t \) reflects the system’s reliability objective and is shaped to encourage selections that meet quality and latency constraints. Specifically, the reward \(\mathcal{R}_t^{\mathcal{C}} \) is computed as:

\vspace{-10pt}
\begin{equation}
\mathcal{R}_t^{\mathcal{C}} = w_1 \cdot S_Q^{(t)} + w_2 \cdot S_L^{(t)}
    \vspace{-5pt}
\end{equation}

where \( S_Q^{(t)} \) and \( S_L^{(t)} \) represent the normalized reconstruction quality and latency scores at time \( t \), and \( w_1, w_2 \in [0,1] \) are weighting coefficients summing one. Latency accounts solely for reconstruction delay, ignoring transmission overhead.

The quality score \( S_Q^{(t)} \) is defined by normalizing the point cloud quality ($Q$) at timestep \( t \) with respect to a reference threshold \( \Theta\):

\vspace{-15pt}
\begin{equation}
S_Q^{(t)} = \min\left(1, \frac{Q_t}{\Theta} \right) 
\vspace{-5pt}
\end{equation}

The latency score \( S_L^{(t)} \) captures how closely the reconstruction latency approaches the system’s real-time threshold \( \Phi\), and is given by:
\vspace{-12pt}
\begin{equation}
    S_L^{(t)} = \max\left(0, 1 - \frac{L_t}{\Phi} \right)  
    \vspace{-5pt}
\end{equation}

This reward formulation enables the agent to receive partial credit for decisions that approach but do not fully satisfy the reliability constraints, thereby supporting progressive learning. The stateless design ensures lightweight, fast decision-making compatible with real-time deployment, while still promoting constraint-aware behavior in dynamic and partially observable environments.

\noindent\underline{\textit{Server selection} --}
The server selection task is formulated as an MDP. 
At each timestep \( t \), after the camera subset is selected, the server agent receives a state \( s_t \) defined as:

\vspace{-5pt}
\begin{equation}
    s_t = \left( \|\mathbf{a}_t\|_1, a_{t-1}^{\mathcal{E}} \right)
    \vspace{-5pt}
\end{equation}
where:
\begin{itemize}
    \item \( \|\mathbf{a}_t\|_1 \) denotes the number of cameras selected at time \( t \),
    \item \( a_{t-1}^{\mathcal{E}} \) is the server selected at the previous timestep.
\end{itemize}

The act of including the number of selected cameras in the server agent’s state allows the agent to condition its server choice on the expected computational load associated with the camera selection decision. 
This coordination between the camera and server selection agents enables more effective adaptation to the dynamic conditions of the system.

The agent then selects an action \( a_t^{\mathcal{E}} \in \mathcal{A}_{\mathcal{E}} \), where \( \mathcal{A}_{\mathcal{E}} \) denotes the set of available servers. The reward for the server agent at timestep \( t \), denoted \( \mathcal{R}_t^{\mathcal{E}} \), is based solely on the end-to-end latency of the 3D reconstruction process, computed as:
\vspace{-5pt}
\begin{equation}
    \mathcal{R}_t^{\mathcal{E}} = S_L^{(t)}   
    \vspace{-5pt}
\end{equation}
where the latency score \( S_L^{(t)} \) is defined as:
\vspace{-2pt}
\begin{equation}
    S_L^{(t)} = \max\left(0, 1 - \frac{L_t}{\Phi}\right)   
    \vspace{-5pt}
\end{equation}


This formulation enables the server agent to learn dynamic server assignment strategies that minimize end-to-end latency under varying server loads and network disruptions.

\vspace{-4pt}
\subsection{Q-learning based solution methodology}
\vspace{-0.05in}

To solve the decision-making problems formulated as MDPs, we employ \textit{tabular Q-learning} for both the camera selection and server selection agents.
Q-learning is a model-free reinforcement learning algorithm that estimates an action-value function \( Q(s, a) \), which represents the expected return of taking action \( a \) in state \( s \) and following the current policy thereafter. At each timestep, the agent observes the current state \( s_t \), selects an action \( a_t \), receives a reward \( r_t \), observes the next state \( s_{t+1} \), and updates the Q-value as:
\vspace{-4pt}

\vspace{-12pt}
\begin{equation}\label{qlearningupdate}
\resizebox{\columnwidth}{!}{$
Q(s_t, a_t) \leftarrow Q(s_t, a_t) + \alpha_t \Big[ r_t + \gamma \max_{a'} Q(s_{t+1}, a') - Q(s_t, a_t) \Big]
$}
\vspace{-5pt}
\end{equation}

This temporal-difference update rule allows the agent to iteratively refine its value estimates without requiring a model of the environment. Q-learning is particularly well-suited to our setting due to the discrete action spaces and its ability to handle delayed rewards, which are common in scenarios where the consequences of decisions (e.g., reconstruction quality or end-to-end latency) may manifest several steps later.

Both agents employ an \(\epsilon\)-greedy strategy for exploration: at each timestep, the agent selects a random action with probability \(\epsilon_t\) and otherwise selects the action that maximizes the current Q-value. This balances exploration of new actions with exploitation of known good actions:

\vspace{-12pt}
\begin{equation}
    a_t =
    \begin{cases}
    \text{random action} & \text{with probability } \epsilon_t \\
    \arg\max_a Q(s_t, a) & \text{with probability } 1 - \epsilon_t
    \end{cases}
    \vspace{-4pt}
\end{equation}

To improve adaptability under non-stationary conditions, we extend Q-learning with adaptive adjustment of both the learning rate \( \alpha_t \) and the exploration rate \( \epsilon_t \), resulting in {\em Adaptive Q-learning}.
When performance degradation is detected 
, both parameters are temporarily increased to encourage faster adaptation:
\vspace{-5pt}
\begin{equation}\label{eq:adapt1}
    \epsilon_{t+1} = \min(\epsilon_t \cdot \eta_{\text{inc}},\ \epsilon_{\max}), \quad \alpha_{t+1} = \min(\alpha_t \cdot \lambda_{\text{inc}},\ \alpha_{\max})
    \vspace{-2pt}
\end{equation}

During steady-state operation, both are gradually decayed to promote convergence:

\vspace{-15pt}
\begin{equation}\label{eq:adapt2}
    \epsilon_{t+1} = \max(\epsilon_t \cdot \eta_{\text{dec}},\ \epsilon_{\min}), \quad \alpha_{t+1} = \max(\alpha_t \cdot \lambda_{\text{dec}},\ \alpha_{\min})  
    \vspace{-2pt}
\end{equation}

Here, \(\eta_{\text{inc}} > 1\), \(\eta_{\text{dec}} < 1\), \(\lambda_{\text{inc}} > 1\), and \(\lambda_{\text{dec}} < 1\) are adaptation factors that control the degree of responsiveness and stabilization. This mechanism enables the agent to remain flexible in the face of environmental volatility while converging efficiently when conditions are stable. In this work, we adopt standard Q-learning based camera selection strategy and adaptive Q-learning based server selection approach. 

\section{Evaluation and Results}
\label{evaluation}



In this section, we evaluate our proposed ad-hoc camera selection and resource management framework through testbed experiments conducted in a realistic Edge-AI environment. We begin by describing the testbed setup, including system components and network design, followed by evaluation metrics, baselines, datasets, and implementation details.
We then present and analyze the experimental results.

\vspace{-8pt}
\subsection{Testbed Setup and Experiment Design}
\begin{figure}[t]
    \centering
    \includegraphics[width=0.85\linewidth]{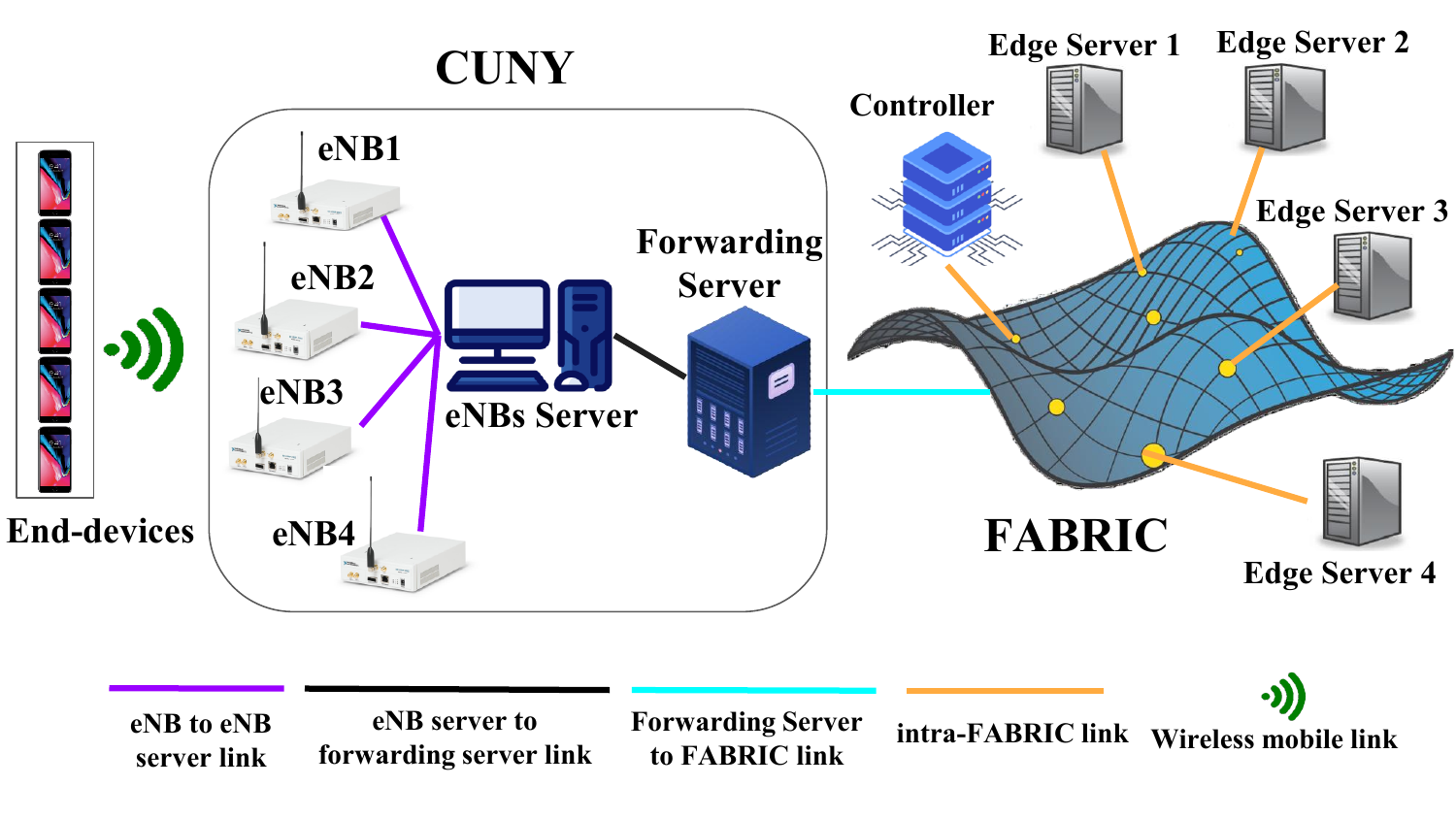}
    \vspace{-0.1in}
    \caption{Smart city testbed implementation}
    \label{fabric}
    \vspace{-0.2in}
\end{figure}

The smart city testbed, illustrated in Fig.~\ref{fabric}, comprises of multiple components:


\subsubsection{End devices} To replicate camera-enabled end-devices, we use 5 Nokia 2.2 smartphones with that has multi-view video dataset already pre-stored. The devices connect to the forwarding server through an LTE network and are synchronized to simultaneously transmit the pre-stored video frames 
    to the forwarding server. 
\subsubsection{Base stations} To enable reliable communication between end devices and forwarding server(s), we deploy eNB-based base stations. These stations handle data transmission over the LTE network, ensuring low-latency connectivity for video frame streaming.
\subsubsection{Forwarding server} To manage data flow between edge devices and end servers, we deploy a forwarding server that receives video streams from the base stations and forwards them to the appropriate edge servers based on controller's decisions. The end devices, base stations, and forwarding servers are implemented within lab hardware at CUNY. 
\subsubsection{End servers} The smart city edge servers are simulated using four virtual servers from FABRIC testbed \cite{fabric-2019}, each equipped with a 16-core CPU (2.8 GHz), 32 GB of RAM, 100 GB of disk space, and an NVIDIA Quadro RTX 6000 GPU for accelerated processing. State-of-the-art 3D reconstruction pipeline are pre-installed on the edge servers.
\subsubsection{Controller} The controller or resource manager is deployed on a machine within FABRIC, equipped with a 20-core CPU (2.8 GHz), 32 GB RAM, and 100 GB disk space. It runs the proposed RL agents, making dynamic decisions on camera and server selection based on system state.

\subsubsection{Network Implementation} 
To create a real-world setting, we establish an LTE network for end-device connectivity to the forwarding server using the srsRAN4G library, enabling a controlled and adaptable LTE environment. The LTE network is supported by multiple eNodeBs, each simulated using USRP-2901 devices with sub-1 GHz antennas. The Evolved Packet Core (EPC), responsible for managing high-level network functions such as user session management, mobility, security, and external data routing, is implemented using \textit{Open5GS}. It is co-located on the same machine as the eNodeBs, creating a compact and integrated smart city environment setup.


\subsubsection{Multi-view 3D reconstruction datasets}
We use our own multi-view dataset \cite{10.1145/3583740.3630267} that
represents different degrees of indoor dynamic scenes, e.g., Pickup, Walk and Handshake. Each scene captures dynamic indoor environments using five synchronized Raspberry Pi cameras, providing long sequence recordings for 3D reconstruction. The scenes are structured with diverse objects, controlled lighting, and a central focal point to ensure consistency. Videos are recorded at 25-30 FPS and are later processed into sequential images for 3D reconstruction. This dataset enables a detailed analysis of 3D reconstruction performance.


\subsubsection{Simulating disruption}
Camera disruption probabilities are simulated over a 4000-frame timeline using a predefined correlation matrix to capture interdependencies. Cameras 1 and 2, and Cameras 3 and 5, are highly correlated, reflecting scenarios where nearby cameras or shared wireless channels fail together (e.g., due to fire, smoke, or interference). Additionally, 10 random bump events—each lasting ~50 frames—are injected across camera groups $\{1,2\}$, $\{3,5\}$, and Camera 4 independently, introducing temporary spikes in failure probability with a threshold of 0.6, producing binary traces (0 = disruption, 1 = normal).
Server-side disruptions are modeled over the same timeline with four independent servers, each having a baseline transmission latency of ~150 ms with small fluctuations. Ten random latency spike events are injected independently, adding 400–1200 ms for ~50 frames per event, emulating temporary server overloads while preserving independent behaviors.

\subsubsection{3D reconstruction Pipeline}
Due to the time-sensitive nature of 3D reconstruction, generating the final dense point cloud using the openMVG/openMVS pipeline creates a latency bottleneck for timely decision-making. On our testbed edge server, this process takes ~5 seconds, exceeding near real-time requirements. However, sparse point clouds generated as intermediate outputs are leveraged to update the RL model’s policy before dense reconstruction completes. In the absence of ground truth, reconstruction quality is quantified by projecting sparse 3D points back onto the original 2D images and measuring the average reprojection error.

\subsubsection{Evaluation metrics} 
We evaluate our RL-based system in terms of observed performance, focusing on reliability, quality, and latency. \textit{Reliability} is the primary metric, defined as the proportion of frames meeting both reconstruction quality and end-to-end latency thresholds (Section~\ref{sec:system model}); \textit{latency} captures the average end-to-end delay per frame from image capture to reconstruction; and \textit{reconstruction quality} is measured using a projection-based metric.

\subsubsection{RL agent training parameters}
We evaluate both fixed and adaptive variants of tabular Q-learning for camera and server selection. All Q-tables are initialized to zero, and training occurs online during system execution. For the camera selection task, the fixed agent uses a learning rate of $\alpha = 0.9$, a discount factor of $\gamma = 0.1$, and a fixed exploration rate of $\epsilon = 0.1$. The adaptive camera agent starts with a base learning rate of $0.5$ and an initial exploration rate of $\epsilon = 1.0$, which decays to a minimum of $0.05$ based on runtime conditions. For server selection, the fixed Q-learning agent is configured with $\alpha = 0.9$, $\gamma = 0.1$, and $\epsilon = 0.1$. The adaptive server agent uses an initial learning rate of $0.3$, an initial exploration rate of $\epsilon = 0.2$, and a discount factor of $\gamma = 0.95$.


\subsubsection{Baseline approaches}
We compare our proposed {Q-learning} and {Adaptive Q-learning} strategies to baselines for camera and server selection, spanning non-learning, heuristic, and learning-based approaches.

\underline{\textit{Camera selection baselines}}: We compare our RL-based strategies against three baselines: (i) \textit{Random}, which uniformly samples a valid camera subset at each timestep as a non-adaptive lower bound; (ii) \textit{Greedy-3}, which always selects the 3-camera subset predicted to yield the highest reconstruction quality, ignoring latency and variability; and (iii) \textit{Epsilon-Greedy Bandit}, which treats camera subset selection as a multi-armed bandit and balances exploration and exploitation via an $\epsilon$-greedy policy.

\underline{\textit{Server selection baselines}}: We compare against two non-learning baselines: (i) \textit{Round-Robin}, which cycles through servers in a fixed order without considering load or latency, serving as a simple deterministic benchmark; and (ii) \textit{Latency-Greedy}, which selects the server with the lowest estimated latency based on an exponentially weighted moving average, adapting to recent trends but without anticipating future variations or accounting for camera quality.

\subsection{Testbed Results and Discussions}
\subsubsection{\textit{Camera selection performance}}
To evaluate different camera selection strategies, we fix the server policy to a \textit{Round-Robin} scheduler so all strategies operate under identical server conditions. Each camera strategy dynamically selects a subset of available cameras at each timestep to optimize quality and responsiveness. Reliability is defined as the percentage of frames meeting all of the following: (1) at least 400 matching points per view, (2) total latency under 3 sec, and (3) reconstruction latency under 1 sec.

\begin{table}[t]
\centering
\caption{Camera selection performance comparison} 
\label{tab:camera_selector_results}
\resizebox{\columnwidth}{!}{%
\begin{tabular}{lcccc}
\toprule
\textbf{Camera Selector} & \textbf{Avg PQ} & \textbf{Avg Recon Lat. (s)} & \textbf{Avg Total Lat. (s)}  & \textbf{Reliability} \\
\midrule
    Q-learning                & 582  & 0.79  & 2.54 &  62.53\% \\
    Greedy-3                  & 542  & 0.67  & 2.45 &  59.92\% \\
    Epsilon-Greedy Bandit     & 443  & 0.71  & 3.45 &  24.69\% \\
    Adaptive Q-learning       & 378  & 0.70  & 2.47 &  36.38\% \\
    Random                    & 386  & 0.69  & 2.26 &  25.05\% \\
\bottomrule
\end{tabular}
}
\vspace{-10pt}
\end{table}

\begin{figure}[t]
    \centering
    \includegraphics[width=1.1\linewidth]{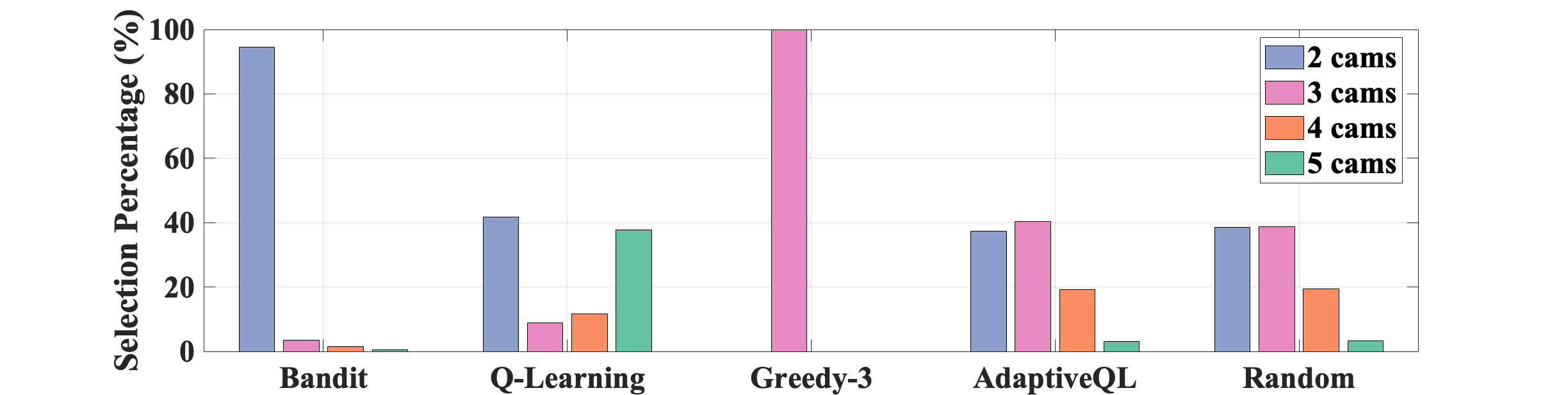}
    \caption{Distribution of selected camera subsets}
    \label{fig:enter-label}
    \vspace{-20pt}
\end{figure}

Tab.~\ref{tab:camera_selector_results} shows that the proposed {\em Q-learning} achieves the highest reliability (62.53\%) and overall point cloud quality, demonstrating its ability to learn policies over time. Interestingly, {\em Greedy-3}, despite being a simple deterministic policy, performs competitively, achieving a reliability of 59.92\% with lower average latency than the Q-learning agent. This suggests that consistently selecting a strong camera subset can serve as a robust baseline in stable environments.
The {\em Epsilon-Greedy Bandit} and {\em Random} strategies perform poorly in terms of both reliability (24.69\% and 25.05\%, respectively) and average point cloud quality. These results reflect their inability to account for delayed feedback and their tendency to make suboptimal choices under variable network conditions. The {\em Adaptive Q-learning} agent, while designed to respond to environmental changes, exhibits instability and lower reliability (36.38\%), likely due to overreaction to short-term fluctuations.
Fig.~\ref{fig:enter-label} visualizes the selection distribution of camera subsets for each strategy. While {\em Q-learning} and {\em Greedy-3} agents concentrate on a limited set of high-quality subsets, the {\em Adaptive Q-learning} and {\em Random} policies show more dispersed behavior. These findings highlight the importance of balancing adaptability and consistency in view selection for real-time 3D reconstruction.

\subsubsection{\textit{Server selection performance}}
To isolate the impact of server selection policies, all experiments use the \textit{Greedy-3} camera selection strategy. Reliability is defined as the percentage of frames meeting all of the following: (1) at least 500 matching points per view, (2) total latency under 3 sec, and (3) reconstruction latency under 1 sec.

\begin{table}[t]
    \centering
    \caption{Server selection performance comparison}
    \label{tab:server_ablation_full}
    \begin{tabular}{lccccccc}
    \toprule
    \textbf{Server Agent} & \textbf{Total Lat. (s)}  & \textbf{Recon Lat. (s)}  & \textbf{Reliability} \\
    \midrule
    Round-Robin         &  3.20  & 0.91  & 31.5\% \\
    Latency-Greedy      &  5.32  & 2.14  & 4.1\% \\
    Q-Learning          &  3.47  & 2.72  & 16.9\% \\
    Adaptive Q-Learning & \textbf{3.62}  & \textbf{2.13} & \textbf{55.0\%} \\
    \bottomrule
    \end{tabular}
    \vspace{-12pt}
\end{table}

\begin{figure}[t]
    \centering
    \includegraphics[width=\linewidth]{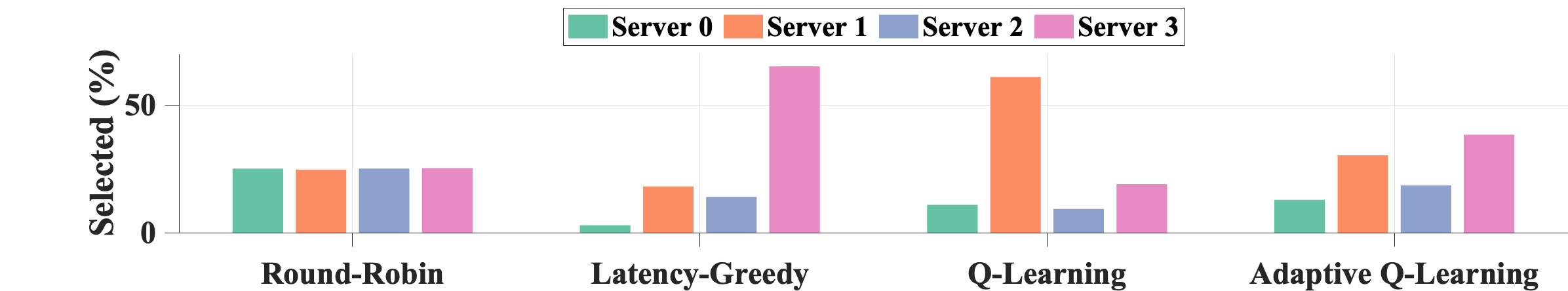}
    \caption{Distribution of selected servers}
    \label{fig:serverselecton_testbed}
    \vspace{-15pt}
\end{figure}

\begin{figure}[t]
    \centering
    \includegraphics[width=\linewidth]{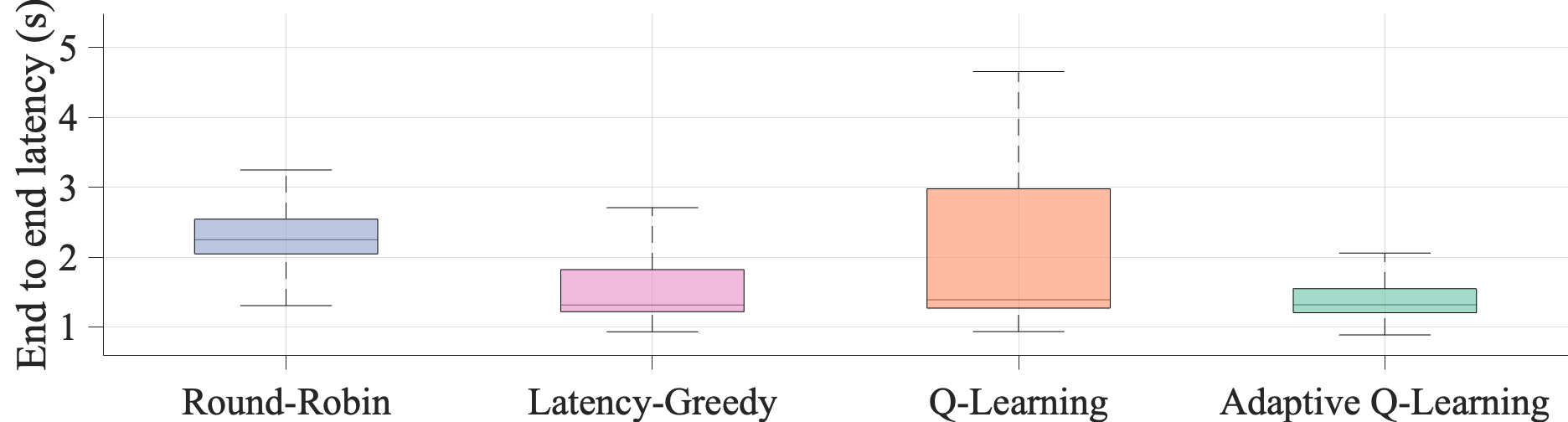}
    \caption{Distribution of end-to-end latency}
    \label{fig:latency_e2e_boxplot}
    \vspace{-18pt}
\end{figure}

Tab.~\ref{tab:server_ablation_full} shows that
{\em Round-Robin} achieves strong performance with a reliability of 31.5\%, outperforming both {\em Latency-Greedy} and {\em Q-learning}. While {\em Q-learning} is designed to adapt based on feedback, it suffers from delayed response to performance degradation and tends to repeatedly select servers that were previously fast, leading to overload. In contrast, the fixed {\em Round-Robin} policy maintains a balanced server distribution — assigning approximately 25\% of frames to each server, which results in more stable latency under dynamic conditions. 
The uniform selection pattern is clearly observable in the server selection results (Fig.~\ref{fig:serverselecton_testbed}). This selection behavior also influences system latency, which varies across the four strategies as shown in Fig.~\ref{fig:latency_e2e_boxplot}.
{\em Round-Robin} exhibits the highest median latency and widest inter-quartile range (IQR) among the more stable strategies, indicating its consistent but non-optimized performance. In contrast, both  {\em Latency-Greedy} and {\em Adaptive Q-learning} show lower medians and narrower IQRs, suggesting more efficient server choices under typical conditions. However, the standard {\em Q-learning} agent displays a much wider spread, with a long box and numerous high-latency outliers. This indicates instability and delayed responsiveness to changing network conditions. 
These results highlight the importance of balancing adaptiveness with robustness in server selection policies.
All evaluation related codes and data are available through Github~\cite{git-edge-cnsm}.
\section{Conclusions}
\label{sec:conclusions}
Latency-sensitive and edge-native multi-view 3D reconstruction is crucial for mission-critical applications, where meeting strict latency and quality constraints enables effective decision-making, often in the presence of disruptions.
We present an RL-based framework that dynamically selects camera subsets and allocates server resources based on observed conditions. The method learns adaptive policies that enhance responsiveness and reliability under unpredictable disruptions. Evaluation in simulated and physical environments shows substantial gains: the camera selection agent achieves up to 15\% higher reliability than random selection and 2\% over Greedy-3, while the adaptive server agent improves up to 24\% over round-robin and 50\% over latency-greedy baselines. These results demonstrate RL’s potential for robust, disruption-aware decision-making in latency-sensitive edge systems. 
\bibliographystyle{ieeetr}
\bibliography{refs}

\end{document}